\documentclass[10pt,twocolumn,letterpaper]{article}

\usepackage{cvpr}              
\usepackage{amsmath}
\usepackage{mathtools}
\usepackage{graphicx}
\usepackage{multirow}
\usepackage{array} 

\usepackage{lipsum} 

\definecolor{chrxcolor}{RGB}{1, 140, 116}

\definecolor{cvprblue}{rgb}{0.21,0.49,0.74}
\usepackage[pagebackref,breaklinks,colorlinks,allcolors=cvprblue]{hyperref}

\usepackage{pifont}
\usepackage{xcolor}
\usepackage{hyperref}

\hypersetup{
    colorlinks=true,
    urlcolor=magenta, 
}
\newcommand{\cmark}{\ding{51}}%
\newcommand{\xmark}{\ding{55}}%

\def\paperID{5920} 
\def\confName{CVPR}
\def\confYear{2026}
\title{MM-OVSeg: Multimodal Optical–SAR Fusion for Open-Vocabulary Segmentation in Remote Sensing}

\author{Yimin Wei$^{1,2,*}$ \quad Aoran Xiao$^{2,*}$ \quad Hongruixuan Chen$^{1,2}$ \quad Junshi Xia$^2$ \quad Naoto Yokoya$^{1,2,\dagger}$\\
$^1$The University of Tokyo \quad $^2$RIKEN AIP\\
{\tt\small 4959184626@edu.k.u-tokyo.ac.jp, \{xiaoaoran94, Qschrx\}@gmail.com,}\\ 
{\tt\small junshi.xia@riken.jp, yokoya@k.u-tokyo.ac.jp}\\
{\small $^*$Equal contribution, $^{\dagger}$Corresponding author}
}

\begin{document}
\maketitle
\begin{abstract}
Open-vocabulary segmentation enables pixel-level recognition from an open set of textual categories, allowing generalization beyond fixed classes. Despite great potential in remote sensing, progress in this area remains largely limited to clear-sky optical data and struggles under cloudy or haze-contaminated conditions. We present MM-OVSeg, a multimodal Optical–SAR fusion framework for resilient open-vocabulary segmentation under adverse weather conditions. MM-OVSeg leverages the complementary strengths of the two modalities—optical imagery provides rich spectral semantics, while synthetic aperture radar (SAR) offers cloud-penetrating structural cues. To address the cross-modal domain gap and the limited dense prediction capability of current vision–language models, we propose two key designs: a cross-modal unification process for multi-sensor representation alignment, and a dual-encoder fusion module that integrates hierarchical features from multiple vision foundation models for text-aligned multimodal segmentation. Extensive experiments demonstrate that MM-OVSeg achieves superior robustness and generalization across diverse cloud conditions. The source dataset and code are available at \href{https://github.com/Jimmyxichen/MM-OVSeg}{\texttt{https://github.com/Jimmyxichen/MM-OVSeg}}.
\end{abstract}


    
\section{Introduction}
\label{sec:intro}

Open-vocabulary segmentation (OVS) aims to assign semantic labels to image regions from an open-ended set of textual categories, enabling recognition beyond the classes observed during training. In remote sensing (RS), OVS is particularly valuable for flexible and scalable land-cover understanding across diverse geographical regions, eliminating the need for exhaustive pixel-level annotations and fixed predefined class sets. This capability promotes improved generalization and adaptability to novel or fine-grained categories commonly encountered in RS imagery.

\begin{figure}[t]
    \centering
    \includegraphics[width=\linewidth]{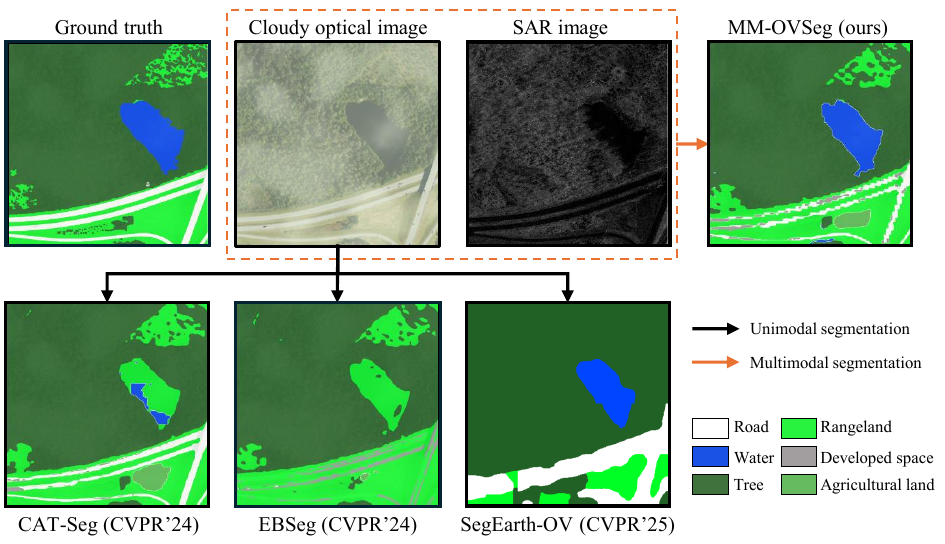}
    \caption{Existing unimodal OVS methods fail in cloudy environments due to severely degraded optical inputs. By incorporating SAR, which penetrates clouds and haze, MM-OVSeg produces significantly more accurate and consistent segmentation results.}
    \label{fig:motivation}
\end{figure}

Despite recent progress, existing OVS studies in the RS domain~\cite{cao2024open,shan2024open,li2025exploring} remain largely confined to optical RGB data, typically assuming clean, clear-sky imagery. In real-world scenarios, however, RS observations are frequently affected by cloud or haze contamination. Current OVS methods struggle under these low-visibility conditions (Figure~\ref{fig:motivation}), thereby limiting their utility for time-sensitive applications such as disaster response and hindering long-term monitoring tasks that require consistent and reliable earth observation and scene understanding.

In this work, we investigate OVS based on the fusion of optical and synthetic aperture radar (SAR) modalities. Leveraging both modalities offers complementary advantages, as optical images provide rich spectral and semantic cues, while SAR data penetrate clouds and capture structural information, enabling robust scene understanding under cloudy or adverse weather conditions.

However, integrating SAR into an open-vocabulary framework is a profoundly challenging and unsolved problem, presenting two key obstacles.  Firstly, vision foundation models (VFMs) are primarily trained on RGB imagery, whereas SAR exhibits distinct backscattering properties and texture patterns, creating a substantial domain gap between RGB and SAR representations. Secondly, vision–language models (VLMs) such as CLIP \cite{radford2021learning} and ALIGN \cite{jia2021scaling} are trained with image-level supervision, limiting their capacity to generate accurate dense predictions for segmentation. This issue is exacerbated for SAR imagery, where domain discrepancies further weaken spatial correspondence between visual and textual representations. Consequently, effectively encoding SAR data in a manner compatible with text-aligned RGB representations, while achieving robust multimodal fusion for OVS remains an open challenge.

In this work, we propose MM-OVSeg, a multimodal Optical–SAR fusion framework for robust open-vocabulary segmentation. The framework comprises two key components designed to tackle the challenges above. First, a Cross-Modal Unification (CMU) process leverages paired RGB–SAR data for cross-modal distillation, aligning SAR embeddings with RGB embeddings distilled from VFMs to establish a shared representation space that enables VFMs to effectively leverage SAR cues. Second, a Dual-Encoder Fusion (DEF) module integrates the CLIP \cite{radford2021learning} encoder (for global semantics) and the DINO \cite{caron2021emerging} encoder (for dense local representations), extracting complementary RGB and SAR features that are fused and aligned with the CLIP text encoder to enable accurate open-vocabulary segmentation. Together, MM-OVSeg provides a unified and resilient multimodal framework for generalizable scene understanding in real-world RS environments.

 Our main contributions are twofold: \textit{First}, we introduce the problem of OVS under cloudy conditions in RS, highlighting the importance of Optical–SAR fusion for robust scene understanding beyond RGB-only imagery. \textit{Second}, we propose MM-OVSeg, a multimodal OVS framework featuring a CMU process for aligning SAR embeddings with RGB-based vision foundation models, and a DEF module for integrating global (CLIP) and local (DINO) features into a unified text-aligned space. Extensive experiments on multiple datasets demonstrate that MM-OVSeg achieves superior accuracy, robustness, and generalization under diverse and cloudy conditions.

\section{Related Work}
\label{sec:formatting}

\subsection{Open-Vocabulary Segmentation}

OVS has gained significant traction in recent years. Early work such as the Open Vocabulary Parsing Network \cite{zhao2017open} explored joint pixel–word embedding spaces. More recent methods \cite{li2022language, zhou2022extract, ghiasi2022scaling, liang2023open, cho2024catseg, shan2024open, li2025fgaseg} leverage large-scale vision–language models, including CLIP \cite{radford2021learning} and ALIGN \cite{jia2021scaling}, to associate arbitrary textual concepts with visual regions. CLIP-based OVS approaches fall into two-stage and one-stage categories. Two-stage methods \cite{ding2022decoupling, liang2023open, xu2022simple} generate category-agnostic region proposals and classify them using CLIP. OVSeg \cite{liang2023open} fine-tunes CLIP with region–text pairs, while MaskCLIP \cite{ding2022open} uses CLIP self-attention for proposal refinement. However, dependence on proposal generators trained with limited annotations often limits their generalization. One-stage models \cite{yu2023convolutions, xu2023side, xie2024sed} avoid external proposals and perform segmentation and recognition jointly. CAT-Seg \cite{cho2024catseg} uses similarity matrices as pseudo-masks, and EBSeg leverages a frozen SAM \cite{kirillov2023segment} image encoder to complement the spatial information missing from CLIP. Despite these advances, existing OVS methods focus on natural images and do not address the unique challenges posed by remote sensing imagery.

\subsection{Open-Vocabulary Segmentation in RS}

Recently, OVS has been extended from natural to RS imagery due to its scalability for land-cover understanding. Current CLIP-based OVS methods in RS can be broadly categorized as training-free or training-required.
Training-free methods leverage CLIP’s inherent localization ability with minimal architectural changes. For instance, SegEarth-OV \cite{li2025segearth} introduces a feature-upsampling module that enhances low-resolution CLIP features while preserving semantic consistency.
Training-required methods, in contrast, learn from labeled base classes to enhance domain adaptation and local detail. Cao et al. \cite{cao2024open} proposed a rotation-aggregative similarity module to handle object rotation and scale variation, while Ye et al. \cite{ye2025towards} and Li et al. \cite{li2025exploring} combined CLIP with a DINO encoder to extract remote-sensing-specific local cues. However, all these methods focus on a single modality, whereas our MM-OVSeg is the first framework for multimodal OVS fusion in RS.

\begin{figure*}[ht]
\centering

\includegraphics[width=\textwidth]{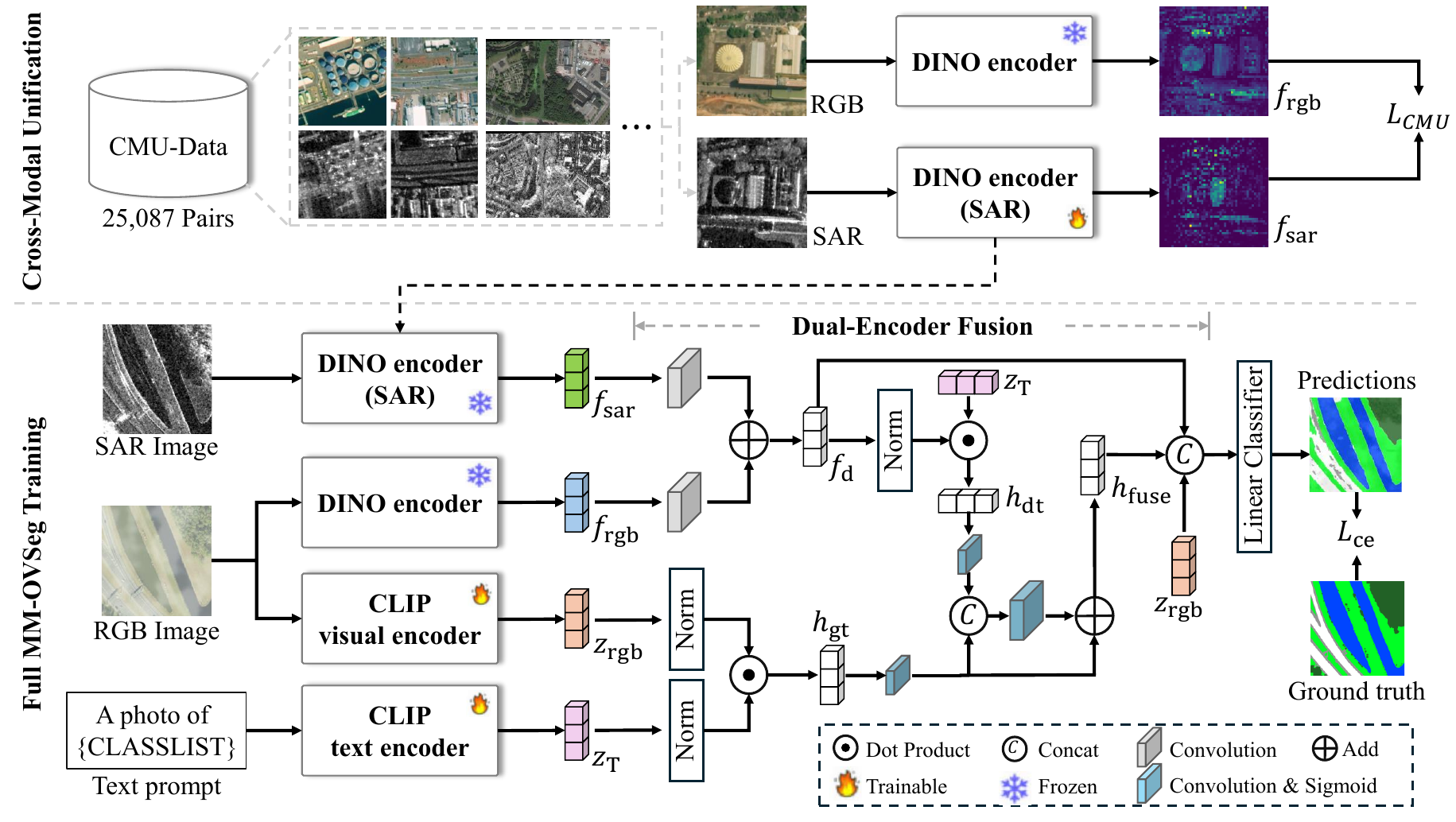}
\caption{Overall optimization framework of MM-OVSeg. The training pipeline consists of two stages. (1) In the Cross-Modal Unification stage, the SAR DINO encoder is trained to align SAR features with the fixed RGB DINO features using the CMU-Data collection of 25,087 RGB and SAR image pairs. (2) In the full MM-OVSeg training stage, the model jointly processes optical and SAR inputs for multimodal open-vocabulary segmentation. The Dual-Encoder Fusion module integrates RGB and SAR dense features and aligns them with CLIP text embeddings, after which a linear classifier predicts the final segmentation map.}
\label{fig:fig2}
\end{figure*}

\subsection{Optical-SAR Integration}
SAR and optical imagery provide complementary cues for Earth observation \cite{xiao2025foundation, wei2026sarlang, Chen2025Bright}. Optical sensors capture spectral and textural information critical for semantic understanding, while SAR penetrates clouds and haze and encodes geometric backscattering, ensuring robustness under adverse conditions.
Fusion of these heterogeneous modalities has been studied at three main levels: pixel, feature, and decision.
Early pixel-level methods combined raw intensity values through techniques such as high-pass filtering, principal component analysis, or Gram–Schmidt transformation \cite{otukei2015fusion, yan2020fusion, kulkarni2020pixel}, yet they were largely application-oriented and limited by modality misalignment.
Feature-level approaches extract modality-specific representations and integrate them via concatenation, attention, or cross-modal adaptation \cite{liu2015pca,garzelli2002wavelet, scheibenreif2022self,yuan2023bridging,liu2023sifnet}, while decision-level fusion combines classifier outputs through probabilistic or evidential rules \cite{waske2008classifying, hughes2020deep}.
Recent deep networks \cite{li2022mcanet,wu2023crofusenet,he2022swin,roy2023multimodal,prakash2021multi,cao2022swin} improve spatial–semantic fusion through attention or transformer modules.
However, existing SAR–optical segmentation methods largely remain closed-set, relying on fixed annotated classes and lacking open-vocabulary generalization.
This limitation motivates our MM-OVSeg, which enables cross-modal fusion for more robust OVS in RS.


\section{Method}

\subsection{Problem Definition}
Given a paired set of multimodal RS images, i.e., an optical RGB image
$I \in \mathbb{R}^{H \times W \times 3}$
and a co-registered  SAR image
$S \in \mathbb{R}^{H \times W \times 1},$
together with a collection of training textual class categories
$\mathcal{C}^{\mathrm{train}} = \{ T(n) \}_{n=1}^{N_c},$
where $T(n)$ denotes the textual description of the $n$-th category and $N_c$ is the number of seen classes,
the objective of OVS is to learn a segmentation model
$\mathcal{G}: (I, S, \mathcal{C}^{\mathrm{train}}) \rightarrow Y,$
that predicts a pixel-wise semantic map
$Y \in \{1, 2, \dots, N_c\}^{H \times W},$
in which each pixel is assigned to the most relevant textual category.

Unlike conventional closed-set segmentation, the label space in OVS is open. At inference time, the model is evaluated on an extended category set
$\mathcal{C}^{\mathrm{test}} = \mathcal{C}^{\mathrm{train}} \cup \mathcal{C}^{\mathrm{novel}},$
where $\mathcal{C}^{\mathrm{novel}}$ represents unseen or novel classes that do not appear during training, i.e.,
$\mathcal{C}^{\mathrm{train}} \cap \mathcal{C}^{\mathrm{novel}} = \emptyset$.
Thus, the model must generalize from seen to unseen categories by aligning visual features with textual semantics in a shared embedding space.

In the multimodal setting, this challenge is further compounded by the heterogeneous nature of the input modalities. The model must effectively fuse optical and SAR features (capturing spectral semantics and structural information, respectively) to produce reliable segmentations under diverse conditions such as cloud or haze contamination. The subsequent subsections detail the proposed framework for achieving this multimodal text–visual alignment.

\subsection{MM-OVSeg}\label{sec:32}


\subsubsection{Overview}
The overall framework of MM-OVSeg is depicted in Figure \ref{fig:fig2}. The model integrates multiple pretrained and learned components to achieve multimodal open-vocabulary segmentation. Specifically, a pretrained CLIP \cite{radford2021learning} model provides a visual encoder $\Phi_V$ and a text encoder $\Phi_T$ for aligning the global visual features of optical RGB images with textual representations. In parallel, a pretrained DINO encoder $\mathcal{F}_{\text{rgb}}$ extracts dense local features from RGB images, while a SAR-specific DINO encoder $\mathcal{F}_{\text{sar}}$ is tuned to produce dense features for SAR images. The RGB and SAR dense features are fused and projected into the CLIP visual feature space, enabling multimodal alignment with the CLIP text embeddings for open-vocabulary prediction.

The training pipeline consists of two stages, as shown in Figure~\ref{fig:fig2}. 
First, the SAR DINO encoder $\mathcal{F}_{\text{sar}}$ is trained through the Cross-Modal Unification (CMU) process to align SAR representations with the fixed RGB DINO features. Then,  the full MM-OVSeg model is trained for joint optical–SAR segmentation, where the Dual-Encoder Fusion (DEF) module performs multimodal feature integration. The following subsections describe them in detail.

\subsubsection{Cross-Modal Unification} \label{sec:CMU}

We employ DINO as the backbone for extracting dense visual features that support segmentation. While pretrained on large-scale RGB corpora, DINO does not directly generalize to SAR imagery, whose microwave backscattering differs substantially from optical texture statistics. As a result, despite its robustness to cloud, haze, and illumination variations, exploiting SAR requires training DINO on this modality.  However, collecting a large-scale SAR corpus on the scale of DINO’s original RGB training set is unrealistic.

To address this challenge, inspired by ImageBind \cite{girdhar2023imagebind}, we unify SAR and RGB embeddings using unlabeled, co-registered RGB–SAR pairs. As shown in Fig. \ref{fig:fig2}(a), each RGB image is encoded by the frozen RGB DINO encoder $\mathcal{F}_{\text{rgb}}$ to obtain $f_{\text{rgb}}$, while its SAR counterpart is processed by the learnable encoder $\mathcal{F}_{\text{sar}}$ to produce $f_{\text{sar}}$. Cross-modal alignment is then optimized via an InfoNCE contrastive loss \cite{oord2018representation} in an unsupervised manner:
$$
L_{\mathrm{CMU}} = -\log \frac{\exp(f_{\text{sar}} f_{\text{rgb}}^{+} / \tau)}
             {\exp(f_{\text{sar}} f_{\text{rgb}}^{+} / \tau) + 
              \sum_{j=1}^{N} \exp(f_{\text{sar}} f_{\text{rgb}}^{-j} / \tau)}
$$
where $f_{\text{rgb}}^{+}$ is the paired RGB embedding, $f_{\text{rgb}}^{-j}$ are negative embeddings from other RGB samples, and $\tau$ is a temperature scalar. Both encoders employ a ViT-B/16 backbone \cite{dosovitskiy2020image}. We extract multi-scale features from the 4th, 8th, and 12th transformer blocks and average their contrastive losses. 
 
To facilitate effective cross-modal learning, we curate CMU-Data, a dataset of 25,087 aligned RGB–SAR pairs collected from SpaceNet6 \cite{shermeyer2020spacenet} and DFC2023 \cite{persello20232023}, with spatial resolutions ranging from 0.5 m to 3 m. Random translation, flipping, scaling, and rotation are used for data augmentation for paired data.

Although one could analogously train a CLIP-style visual encoder for SAR, we find this unnecessary in practice. The pretrained CLIP visual encoder captures global semantic cues (scene layout, object co-occurrence, and contextual relations) that remain largely invariant across optical and SAR modalities. Thus, training an additional CLIP encoder for SAR introduces significant overhead without meaningful gains. We provide further discussion in the appendix.

\subsubsection{Dual-Encoder Fusion}

The second stage of MM-OVSeg trains the full multimodal OVS framework. As shown in Fig. \ref{fig:fig2}(b), the pretrained RGB DINO encoder $\mathcal{F}_{\text{rgb}}$ and the CMU-aligned SAR DINO encoder $\mathcal{F}_{\text{sar}}$ are frozen, while the CLIP visual and text encoders $(\Phi_V, \Phi_T)$ remain trainable. The key component in this stage is the \textit{Dual-Encoder Fusion (DEF)} module, which integrates CLIP and DINO to extract complementary RGB–SAR cues and align them with textual semantics.








\textbf{Multimodal dense feature aggregation.}
Given paired inputs $(I_{\text{rgb}}, I_{\text{sar}})$, dense features are extracted from the 4th, 8th, and 12th transformer blocks of the ViT-B/16 backbone:
$$f_{\text{rgb}}^{i}=\mathcal{F}_{\text{rgb}}^{i}(I_{\text{rgb}}), \qquad
f_{\text{sar}}^{i}=\mathcal{F}_{\text{sar}}^{i}(I_{\text{sar}}), \quad i\in\{1,2,3\},$$
Each feature map is projected to a unified dimension by a block-specific convolution layer $\sigma_i(\cdot)$. The RGB and SAR features are then fused via element-wise addition:
$$f_d^{i} = \sigma_i(f_{\text{rgb}}^{i}) + \sigma_i(f_{\text{sar}}^{i}).$$
This multimodal representation incorporates spectral–textural cues from RGB and structural backscatter from SAR.

\begin{table*}[t]
    \centering
    \setlength{\tabcolsep}{15pt}  
    \begin{tabular}{c|l|cccc}
    \toprule
       Index & Datasets (\textit{train} → \textit{test})  & weather & cloud cover & cloud type & generalization \\
    \midrule
       \ding{172} & PIE-cloud → PIE-cloud & cloudy & varied & synthetic & intra-domain\\
       \ding{173} & DDHR-SK → DDHR-SK & cloudy & varied & synthetic & intra-domain\\
       \ding{174} & OEM-thick → OEM-thick & cloudy & thick & synthetic & intra-domain\\
       \ding{175} & OEM-thin → OEM-thin & cloudy & thin & synthetic & intra-domain\\
       \ding{176} & PIE-clean → PIE-clean & clear sky & \textit{none} & \textit{none} & intra-domain\\
       \ding{177} & DDHR-SK → DDHR-CH & cloudy & varied & synthetic & cross-domain \\
    \bottomrule
    \end{tabular}
    \caption{Evaluation settings for MM-OVSeg. The experiments cover clear sky and cloudy weather, synthetic cloud cover with different opacity levels (thin or thick or varied), and both intra-domain and cross-domain generalization scenarios.}
    \label{tab:datasets}
\end{table*}

\textbf{Text–visual alignment.}
For the same RGB image and corresponding text prompt $T$, CLIP produces global visual and textual embeddings:
$z_{\text{rgb}} = \Phi_V(I_{\text{rgb}}), \quad
z_T = \Phi_T(T),$
where the text prompt follows the template “a photo of \{CLASSLIST\}” for all training categories $\mathcal{C}^{\mathrm{train}}$.
Dense visual–text and global visual–text similarities are computed using cosine similarity:
$h_{dt}^i = f_d^i \cdot z_T, \quad
h_{gt} = z_{\text{rgb}} \cdot z_T.$
These similarity maps are transformed by a $7{\times}7$ convolution and sigmoid activation:
$h_{dt}^i, h_{gt} = \varphi(\sigma_7(h_{dt}^i)), \varphi(\sigma_7(h_{gt}))$.
To jointly leverage dense and global alignment, DEF fuses them through concatenation and residual enhancement:
$$ h_{\text{fuse}}^i = \varphi(\sigma_7([h_{dt}^i; h_{gt}])) + h_{gt}$$  
The residual connection preserves the generalist semantic structure encoded by CLIP and mitigates feature drift during multimodal training.

Following an FPN-style decoder \cite{lin2017feature}, fused features $h_{\text{fuse}}^{i}$ across blocks are bilinearly upsampled and concatenated with corresponding DINO and CLIP features. A linear classifier is applied to generate pixel-wise predictions.

Training is supervised using standard cross-entropy loss $L_{ce}$.
At inference, text prompts for all categories in $\mathcal{C}^{\mathrm{test}}$ are injected into the model, enabling open-vocabulary segmentation across both seen and unseen classes.

\section{Experiments}

\begin{table*}[t] 
    \setlength{\tabcolsep}{12pt}
    \centering
    \begin{tabular}{l|c|cccccc|c}
    \toprule
        Method & Publication & \ding{172} & \ding{173} & \ding{174} & \ding{175} & \ding{176} & \ding{177} & Mean \\
    \midrule
        CAT-Seg \cite{cho2024catseg} & CVPR'24 & 54.5 & 54.2 & 33.8 & 29.5  & 55.8 & 27.8 & 42.6\\

        EBSeg \cite{shan2024open} & CVPR'24 & 50.8 & 51.1 & 27.2 & 25.6 & 51.0 & 26.7 & 38.7\\

        GSNet \cite{ye2025towards} & AAAI'25 & 57.0 & 55.0 & 35.2 & 37.0 & 57.2 & 32.4 & 45.6\\

       SegEarth-OV \cite{li2025segearth} & CVPR'25 & 45.1 & 17.6 & 28.9 & 18.5 & 51.8 & 24.2 & 31.0\\


        FGAseg \cite{li2025fgaseg} & arXiv'25 & 51.6 & 51.6 & 26.0 & 32.8 & 52.1 & 40.6 & 42.5\\
        \textbf{MM-OVSeg (ours)} & -- & \textbf{57.7} & \textbf{73.1} & 
        \textbf{36.6} &
         \textbf{40.2} &  \textbf{59.7} & 
         \textbf{42.6} &
         \textbf{51.7} \\
    \bottomrule
    \end{tabular}
    \caption{Comparison of OVS methods across all evaluation settings defined in Table \ref{tab:datasets}. The table reports mIoU scores for each setting and the overall mean. Settings correspond to:
    \ding{172}: PIE-cloud→PIE-cloud; \ding{173}: DDHR-SK→DDHR-SK; \ding{174}: OEM-thick→OEM-thick; \ding{175}: OEM-thin→OEM-thin; \ding{176}: PIE-clean→PIE-clean; \ding{177}: DDHR-SK→DDHR-CH. MM-OVSeg achieves the highest accuracy in all settings and obtains the best overall mean score, demonstrating strong robustness under cloudy conditions and superior cross-domain generalization.}
    \label{tab:table2}
\end{table*}

\begin{figure*}[t]
\centering
\includegraphics[width=0.8\linewidth]{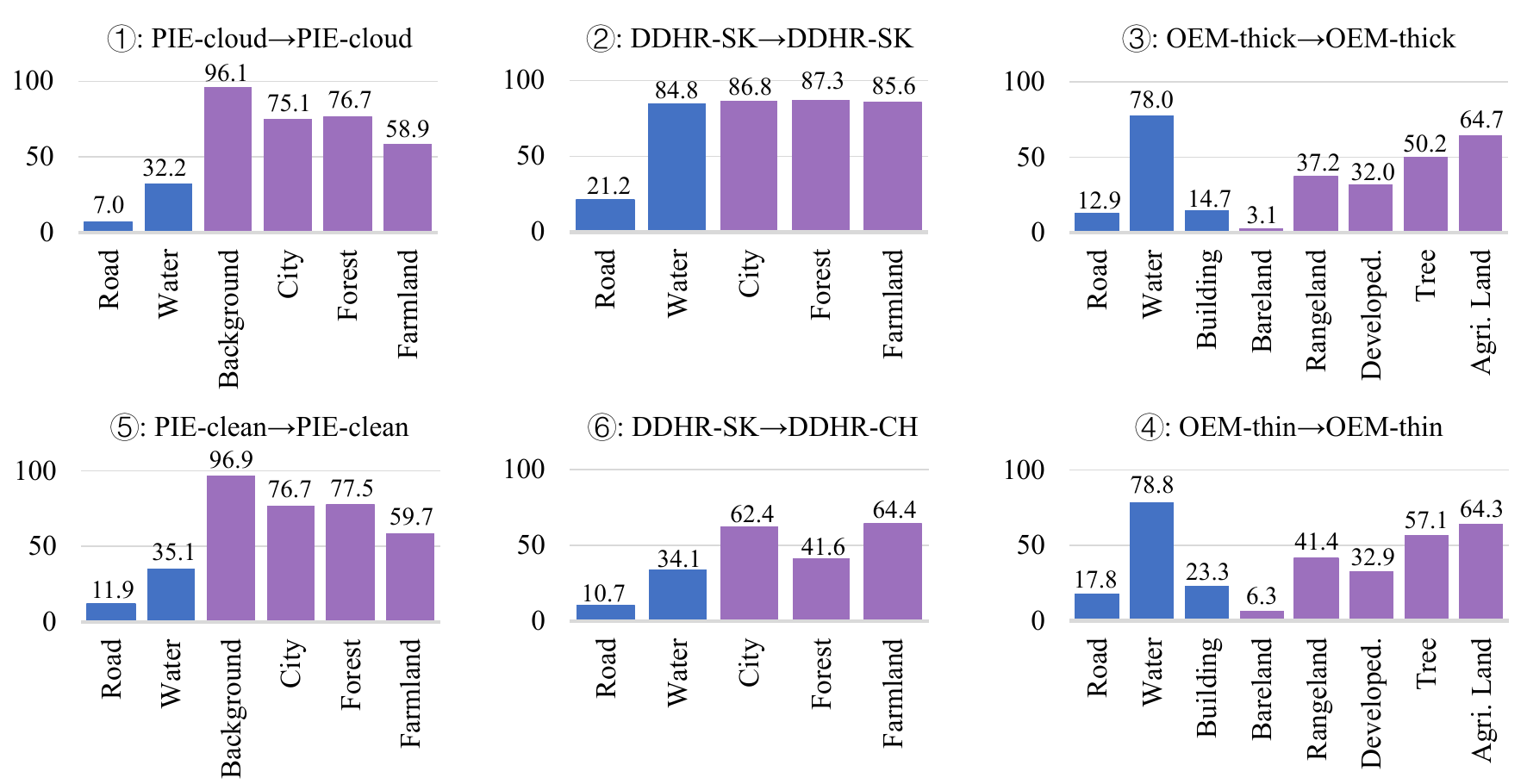}
\caption{IoU performance for each individual class under the six evaluation settings defined in Table \ref{tab:datasets}. \textcolor{purple}{Purple} bars and \textcolor{blue}{blue} bars represent \textit{seen} and \textit{unseen} classes, respectively.}
\label{fig:figzhu}
\end{figure*}

\subsection{Datasets and Settings}

\subsubsection{Experimental Setups}
We comprehensively evaluate MM-OVSeg across multiple multimodal RS datasets under diverse weather and domain conditions. Our evaluation encompasses (1) \textbf{clear-sky vs. cloudy weather}, (2) \textbf{synthetic cloud} cover with varying opacity (\textbf{thin vs. thick vs. varied}), and (3) \textbf{intra-domain and cross-domain} generalization.

\textbf{OpenEarthMap-SAR} \cite{xia2025openearthmap} comprises 1.5 million segmented tiles (1024 × 1024 pixels) collected from aerial and satellite imagery over 35 regions in Japan, France, and the United States, with a ground sampling distance (GSD) of 0.15–0.5 m. Eight land-cover categories are annotated. Following the official split, we use 4,333 RGB–SAR image pairs for training and 490 pairs for testing.
Since the dataset contains only clear-sky imagery, we apply SatelliteCloudGenerator \cite{rs15174138} to synthesize cloud cover, producing two variants: \textbf{OEM-thin} and \textbf{OEM-thick}.
Training classes include \textit{Bareland, Rangeland, Developed space, Tree}, and \textit{Agricultural land}, while testing additionally includes novel classes \textit{Road, Water}, and \textit{Building}.

\textbf{PIE-RGB-SAR} \cite{ZHANG2024574} contains multimodal image pairs from the Pearl River Delta, China, with RGB images sourced from Google Satellite and SAR images from the GF-3 satellite’s ultra-fine stripe mode. The RGB and SAR images have an approximate GSD of 0.5 m and 3 m, respectively. All pairs are cropped to 256 × 256 pixels, yielding 4,865 samples (2,433 for training and 2,432 for testing). Two tracks are officially provided: \textbf{PIE-clean} (cloud-free) and \textbf{PIE-cloud} (cloudy). We use training classes \textit{Background, City, Forest}, and \textit{Farmland}, and define \textit{Road} and \textit{Water} as novel test classes.

\textbf{DDHR} \cite{ren2022dual} includes RGB images from the GF-2 satellite, synthetically clouded using the GNU Image Manipulation Program (GIMP) software, paired with SAR images from GF-3 resampled to 1 m resolution. Both modalities are cropped to 256 × 256 pixels.
The dataset contains five subsets; we use two: \textbf{DDHR-SK} (Pohang, South Korea) and \textbf{DDHR-CH} (Xi’an, China).
For DDHR-SK, we follow the official split with 3,087 training and 3,086 validation images. DDHR-CH is used entirely for cross-domain testing.
Five classes are annotated, among which \textit{Forest, City}, and \textit{Farmland} are used for training, and \textit{Road} and \textit{Water} are treated as novel classes in testing.

{\bf Evaluation Protocol.} Table \ref{tab:datasets} summarizes the evaluation setups across the datasets described above.
We report mean Intersection over Union (mIoU) as the primary metric for evaluation of semantic segmentation performance.

\subsubsection{Implementation Details}\label{sec:412}

We implement MM-OVSeg in PyTorch \cite{paszke2019pytorch} using Detectron2 \cite{wu2019detectron2} for the segmentation pipeline. Both CLIP and DINO components use a ViT-B/16 backbone. Pretrained weights are taken from CLIP \cite{radford2021learning} and DINO v1 \cite{caron2021emerging}.
(1) CMU training: We train the SAR DINO encoder with a batch size of 8 using AdamW \cite{loshchilov2017decoupled}. The learning rate is set to $3\times10^{-4}$ with a weight decay of $1\times10^{-4}$. The RGB DINO encoder remains frozen. 
(2) Full MM-OVSeg training:
All newly introduced parameters are initialized randomly. We fine-tune the model for 120k iterations using AdamW with a batch size of 8 and an initial learning rate of $2.5\times10^{-4}$. The CLIP encoders are trained with a smaller learning rate of $2\times10^{-6}$ to preserve their pretrained alignment. All experiments are conducted on a single NVIDIA RTX A100 GPU (80 GB).

\subsubsection{Baselines and Comparison Methods}
Since the proposed MM-OVSeg is the first multimodal fusion framework for OVS in RS, we compare it with six state-of-the-art single-modality OVS models.
These include
1) OVS models in the natural image domain: CAT-Seg \cite{cho2024catseg}, EBSeg \cite{shan2024open}, and FGAseg \cite{li2025fgaseg};
2) OVS models in the RS domain: GSNet \cite{ye2025towards}; and
3) a training-free RS OVS model: SegEarth-OV \cite{li2025segearth}, used as a reference baseline.
\subsection{Main Result}

Table \ref{tab:table2} presents a quantitative comparison between MM-OVSeg and recent state-of-the-art methods across all benchmark datasets. mIoU is reported for each dataset and the overall mean.
Overall, MM-OVSeg achieves the highest average performance, reaching 51.7\% over six benchmarks, while the second-best method, GSNet, obtains 45.6\%. This substantial improvement highlights the effectiveness of the proposed multimodal fusion framework in advancing open-vocabulary segmentation within the remote sensing domain. Moreover, unlike prior methods that only excel either on seen or unseen classes, MM-OVSeg demonstrates consistently strong performance across datasets for both \textit{seen} and \textit{unseen} classes. Please refer to Table \ref{tab:performanceSplit} in the appendix for split seen/unseen class performance comparisons.


Figure~\ref{fig:figzhu} further visualizes IoU performance for each individual class. While MM-OVSeg outperforms all competing methods by large margins, the model still achieves notably higher accuracy on seen classes than on novel unseen categories, underscoring the inherent challenge of OVS in remote sensing—particularly in maintaining robust visual–text alignment. Interestingly, MM-OVSeg attains especially strong performance on the unseen water category. This can be attributed to the characteristically low and homogeneous backscatter of water surfaces in SAR imagery, which provides a reliable cue for discrimination and illustrates the benefit of incorporating SAR into the multimodal framework. 

\begin{figure*}[ht]
\centering
\includegraphics[width=0.86\linewidth]{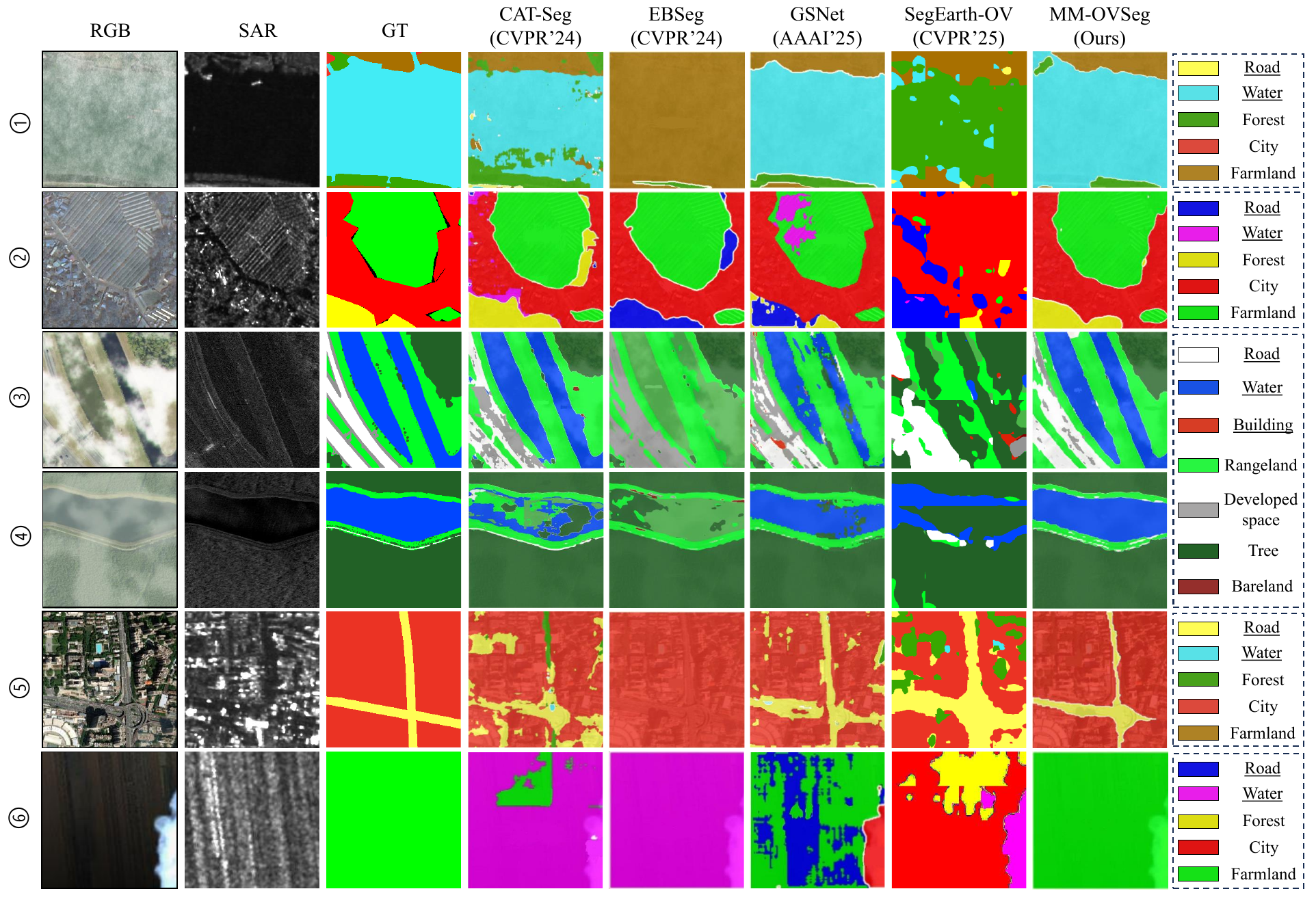}
\caption{Visualization of OVS results. From left to right: input RGB image, input SAR image, ground truth, and segmentation outputs from CAT-Seg, EBSeg, GSNet, SegEarth-OV, and our MM-OVSeg. In the legend, underlined categories represent \textit{unseen} classes and the remaining categories are \textit{seen} classes.}
\label{fig:fig3}
\end{figure*}

\textbf{Cross-weather robustness.} MM-OVSeg delivers consistently strong performance under various cloudy and hazy conditions. Its adaptive fusion mechanism effectively exploits the cloud-penetrating capability of SAR while dynamically weighting RGB and SAR cues according to scene quality. As a result, MM-OVSeg achieves the best performance in all weather scenarios. Notably, even on the clear-sky benchmark (\ding{176}: PIE-clean → PIE-clean), where SAR contributions are less critical, MM-OVSeg still surpasses GSNet by 2.5\% mIoU.

\textbf{Cloud-type and cloud-cover variability.} Across synthetic clouds, including thin, thick, and mixed layers, MM-OVSeg yields stable segmentation accuracy. This indicates that the model learns to exploit complementary spectral and structural cues from the two modalities rather than overfitting to specific cloud patterns, confirming the flexibility of our fusion strategy.

\textbf{Domain generalization.} To evaluate cross-domain robustness, we train on DDHR-SK and test on DDHR-CH (\ding{177}: DDHR-SK→DDHR-CH), which represent distinct geographic regions. As expected, all training-required methods experience performance degradation when compared with intra-domain testing (\ding{173}: DDHR-SK→DDHR-SK), reflecting significant domain discrepancy. Nevertheless, MM-OVSeg maintains a clear margin over all competitors, demonstrating leading adaptability to unseen regions.

\textbf{Qualitative analysis.}
Figure~\ref{fig:fig3} provides visual comparisons. Previous models are highly sensitive to cloud interference: thick clouds often cause misclassification, and even thin haze layers can distort predictions. In contrast, MM-OVSeg produces coherent and accurate segmentations by adaptively leveraging both optical and SAR information. Moreover, the model correctly identifies novel categories that are unseen during training, validating its strong text-feature alignment. Additional qualitative results are provided in the appendix.
\begin{table}[t]
    \renewcommand{\arraystretch}{1.15}
    \resizebox{\linewidth}{!}{%
    \centering
    \begin{tabular}{l|ccccc|c}
    \toprule
       Model Variant & Forest & City & Farmland & Road & Water & mIoU \\
    \midrule
       MM-OVSeg & 87.3 & 86.8 & 85.6 & 21.2 & 84.8 & 73.1 \\
       \textit{w/o} CMU  & 57.2 & 83.7 & 81.2 & 16.8 & 81.4 & 64.1 \\
       \textit{w/o} CMU\&DEF  & 80.0 & 90.3 & 79.0  & 6.8 & 19.1 & 55.0 \\
    \bottomrule
    \end{tabular}}
    \caption{Ablation study of MM-OVSeg on the DDHR-SK→DDHR-SK segmentation task under cloudy conditions. The proposed DEF module enables effective multimodal fusion, substantially improving over the single-modality baseline. In combination with CMU, the full model achieves the best performance, demonstrating that CMU and DEF are complementary.}
    \label{tab:ablation}
\end{table}

\subsection{Ablation Studies}

We conduct ablation experiments to analyze the contribution of each component of MM-OVSeg to open-vocabulary segmentation. All experiments are performed on the DDHR-SK dataset, and results are summarized in Table~\ref{tab:ablation}. The baseline model, which uses only optical data without the proposed Cross-Modal Unification (CMU) or Dual-Encoder Fusion (DEF) modules, achieves 55.0\% mIoU, revealing the limitation of single-modality segmentation. Introducing the DEF module leads to a clear improvement of 9.1\%, confirming the effectiveness of multimodal feature integration between RGB and SAR data. Finally, incorporating both CMU and DEF yields the full MM-OVSeg, which achieves the best performance of 73.1\%. These results demonstrate that CMU and DEF are complementary: CMU effectively aligns SAR features with RGB representations, while DEF fuses them adaptively for robust multimodal segmentation.

\begin{figure}[t]
\centering
\includegraphics[width=\linewidth]{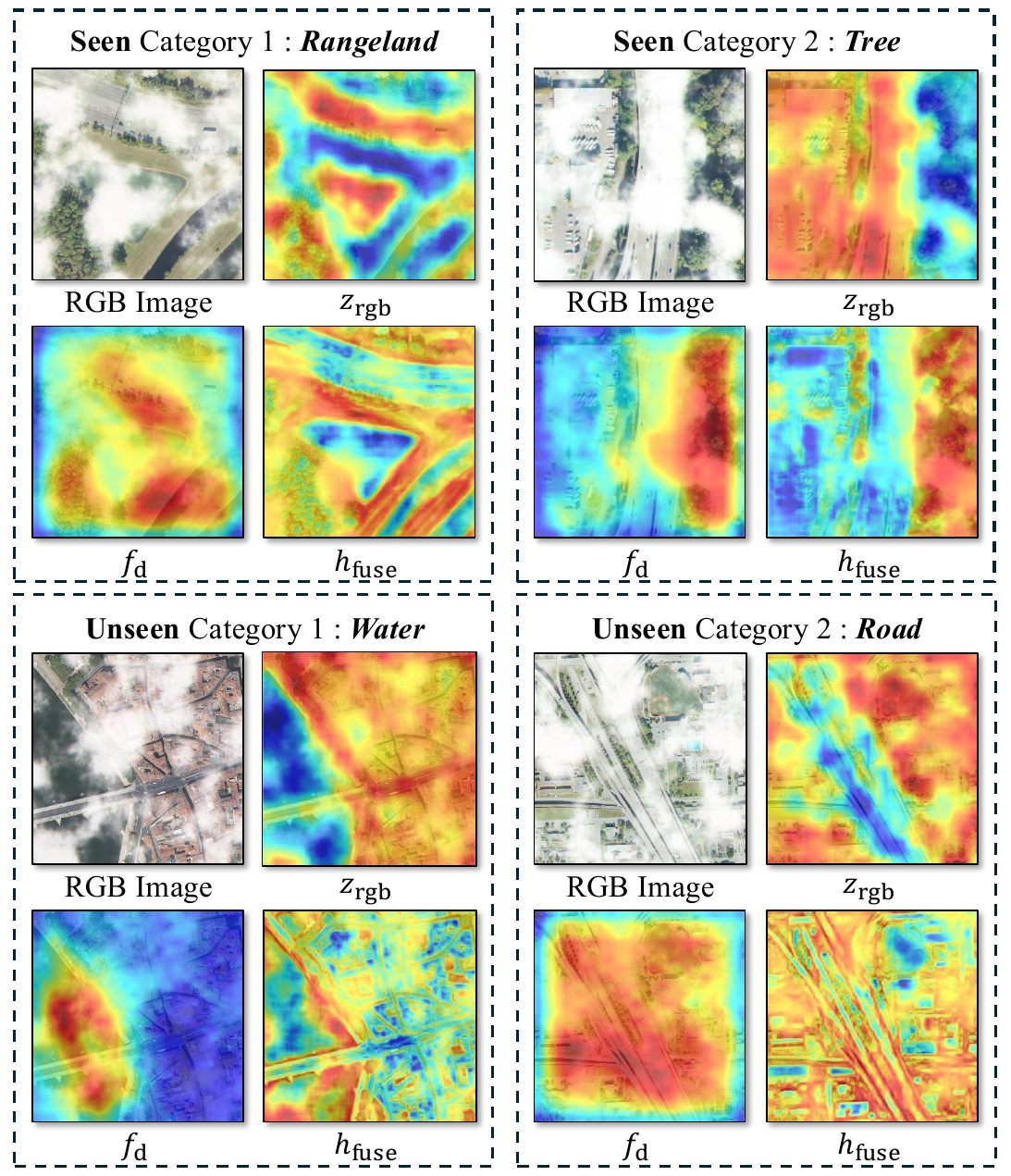}
\caption{Visualization of different stages in multimodal fusion within DEF. DEF produces finer spatial localization and stronger alignment between dense and global visual features and text representations.}
\label{fig:fig4}
\end{figure}

\subsection{Discussions}

\subsubsection{Feature Visualization and Analysis of DEF}

To better understand how DEF integrates representations from different foundation models and modalities, we visualize the intermediate feature maps generated at various stages of the fusion process.
As illustrated in Figure \ref{fig:fig4}, we show (1) the input RGB image for reference, (2) the global visual feature $z_{\text{rgb}}$ extracted by the CLIP visual encoder, (3) the multimodal dense feature $f_d$ obtained from DINO and DINO-SAR, and (4) the final fused representation $h_{\text{fuse}}$, which combines dense RGB and SAR features with global CLIP visual and text embeddings.
We use text prompts corresponding to two seen categories (Rangeland, Tree) and two unseen categories (Water, Road), and visualize their attention heatmaps.

The results reveal several insights.
First, the CLIP visual encoder is relatively less affected by cloud interference than DINO, but its attention maps remain coarse because it primarily captures global semantics.
Second, CLIP and DINO exhibit complementary attention patterns—regions highlighted strongly by one are often suppressed by the other—indicating that their feature spaces encode different but synergistic cues.
Finally, the fused representation $h_{\text{fuse}}$ provides much finer spatial localization and better alignment with text prompts, accurately highlighting both seen and unseen categories.
These visualizations confirm that DEF effectively unifies dense and global representations across modalities, leading to improved semantic correspondence and robustness under cloudy conditions.



\begin{table}[t]
    \setlength{\tabcolsep}{20pt} 
    \centering
    \begin{tabular}{l|c}
    \toprule
        Loss type & mIoU \\
    \midrule
        \textit{baseline} (\textit{w/o} CMU) & 64.1 \\
        MSE & 67.7 \\
        L1 & 69.0 \\
        InfoNCE & \textbf{73.1} \\
    \bottomrule
    \end{tabular}
    \caption{Comparison of loss functions used in the CMU stage.}
    \label{tab:losses}
\end{table}

\subsubsection{Loss Design in CMU}

We analyze the impact of different loss functions used in the CMU stage. Specifically, we compare MSE loss, L1 loss, and InfoNCE loss to evaluate how they affect the feature alignment between SAR and RGB representations in the DINO encoders. The model without CMU is also included as a baseline for reference. As shown in Table \ref{tab:losses}, all three loss functions yield notable improvements over the baseline, confirming the effectiveness of the CMU. Among them, the InfoNCE loss achieves the highest performance, surpassing both MSE and L1 losses by a clear margin.

\section{Conclusion}


We presented MM-OVSeg, the first multimodal Optical–SAR framework for open-vocabulary segmentation (OVS) in remote sensing.
By introducing two complementary modules, namely Cross-Modal Unification (CMU) for aligning SAR representations with RGB features, and Dual-Encoder Fusion (DEF) for integrating dense and global visual features with textual embeddings, MM-OVSeg effectively bridges the modality and semantic gaps that limit existing OVS approaches.
Extensive experiments across diverse weather and domain conditions demonstrate that MM-OVSeg achieves superior robustness, generalization, and segmentation accuracy compared with recent state-of-the-art models.



\noindent\textbf{Acknowledgement.} This work was supported by JST FOREST (Grant No. JPMJFR206S), JST CRONOS (Grant No. JPMJCS25K5), JSPS KAKENHI (Grant No. 24KJ0652), and Next Generation AI Research Center of UTokyo.


{
    \small
    \bibliographystyle{ieeenat_fullname}
    \bibliography{main}
}

\clearpage
\appendix
\clearpage
\setcounter{page}{1}
\maketitlesupplementary

\renewcommand{\thefigure}{A\arabic{figure}}
\renewcommand{\thetable}{A\arabic{table}}
\setcounter{figure}{0}
\setcounter{table}{0}



In this appendix, we present additional experimental results and analyses. Section \ref{sec:implement} provides supplementary training details beyond Section \ref{sec:412} of the main paper. Section \ref{sec:efficiency} reports the efficiency and sustainability comparison. To further demonstrate the generalization capabilities of MM-OVSeg, we provide Section \ref{sec:splitPerformance} with more detailed split performance of seen/unseen classes across six datasets. Section \ref{sec:vitl14} reports results of MM-OVSeg using the ViT-L/14 backbone. Section \ref{sec:sarclip} provides further studies on CMU, including its extension to CLIP-SAR alignment. Section \ref{sec:visual} includes additional qualitative visualizations. 

\section{  Implementation Details} \label{sec:implement}

In Table \ref{tab:implement}, we present more detailed implementations of our MM-OVSeg and other methods, including trainable parameters, batch size and learning rate.

\begin{table}[!htbp]
\centering
\renewcommand{\arraystretch}{1.2}
\resizebox{\linewidth}{!}{
\begin{tabular}{l|c|c|c|c}
\hline
Method & Backbone & Param. & Batch. & Lr \\
\hline
 CAT-Seg \cite{cho2024catseg} & ViT-B/16 & \textbf{25M} & 4 & 2e-4  \\
 FGAseg \cite{li2025fgaseg} & ViT-B/16 & 33M & 8 & 2e-4 \\
 GSNet \cite{ye2025towards} & ViT-B/16 & 29M & 4 & 2e-4 \\
 EBSeg \cite{shan2024open} & ViT-B/16 & 26M & 16 & 1e-4 \\
 MM-OVSeg & ViT-B/16 & 31M & 8 & 2.5e-4  \\
\hline
\end{tabular}}
\caption{Model implementations. The table reports trainable parameters, batch size and learning rate (Lr).}
\label{tab:implement}
\end{table}


\section{  Efficiency and Sustainability Comparison} \label{sec:efficiency}

In Table \ref{tab:efficiency}, we report training/inference latency, estimated carbon emissions \cite{henderson2020towards}, and parameter counts for MM-OVSeg and GSNet on DDHR-SK→DDHR-SK. MM-OVSeg incurs additional cost due to SAR integration and dual encoders, but this overhead is accompanied by improved segmentation performance and robustness under adverse conditions.



\begin{table}[h]
\centering
\renewcommand{\arraystretch}{1.25}
\setlength{\tabcolsep}{3pt}
\resizebox{\linewidth}{!}{
\begin{tabular}{l|cc|cc|c|c}
\hline
 & \multicolumn{2}{c|}{Time (ms)} & \multicolumn{2}{c|}{CO$_2$ (kg)} & \multirow{2}{*}{Parameters} & \multirow{2}{*}{mIoU} \\
 & Train & Infer & Train & Infer & & \\
\hline
CMU module & 343.7  & 31.6  & 1.05 & 0.006 & 85M  & -- \\
DEF module & 377.2 & 46.8 & 1.14 & 0.008 & 92M & -- \\
\hline
\textit{full} MM-OVSeg  & 955.7  & 116.9  & 3.90 & 0.014 & 331M & \textbf{73.1} \\
\hline
GSNet \cite{ye2025towards} & 700.6 & 83.2 & 3.48 & 0.011 & 244M & 55.0\\
\hline
\end{tabular}
}
\caption{Efficiency and sustainability comparison.}
\label{tab:efficiency}
\end{table}




\section{  Seen vs. Unseen Class Evaluation} \label{sec:splitPerformance}

In Table \ref{tab:performanceSplit}, we provide a more detailed breakdown of seen and unseen class performance, supplementing Table \ref{tab:table2} from the main paper. All methods show reduced performance on unseen classes, which is expected in OVS. Prior methods often exhibit imbalanced behavior, performing well on either seen or unseen classes but not both, with inconsistent trends across datasets. In contrast, MM-OVSeg shows more balanced and consistently strong performance on both seen and unseen classes across datasets, indicating improved robustness in the OVS setting.

\begin{table*}[h]
    \renewcommand{\arraystretch}{1.45}
    \setlength{\tabcolsep}{3pt}
    \centering
    \resizebox{\textwidth}{!}{
    \begin{tabular}{>{\centering\arraybackslash}m{2.8cm}|>{\centering\arraybackslash}m{1.7cm}|cc|cc|cc|cc|cc|cc|cc}
    \hline
        \multirow{2}{*}{Method} & \multirow{2}{*}{Publication} & \multicolumn{2}{c|}{\ding{172}} & \multicolumn{2}{c|}{\ding{173}} & \multicolumn{2}{c|}{\ding{174}} & \multicolumn{2}{c|}{\ding{175}} & \multicolumn{2}{c|}{\ding{176}} & \multicolumn{2}{c|}{\ding{177}} & \multicolumn{2}{c}{Mean} \\
        & & Unseen & Seen & Unseen & Seen & Unseen & Seen & Unseen & Seen & Unseen & Seen & Unseen & Seen & Unseen & Seen \\
    \hline
        CAT-Seg \cite{cho2024catseg} & CVPR’24 & 14.9 & 74.2 & 41.1 & 62.9 & 32.2 & 34.6 & 17.0 & 36.9 & 17.1 & 75.1 & 7.5 & 41.5 & \underline{21.6} & 54.2 \\
        EBSeg \cite{shan2024open} & CVPR’24 & ~2.5 & 74.8 & 26.7 & 67.4 & 12.7 & 35.9 & ~5.5 & 37.6 & 0.8 & 76.0 & 8.5 & 38.8 & ~9.5 & 55.1 \\
        GSNet \cite{ye2025towards} & AAAI’25 & 18.6 & \underline{76.1} & 13.0 & \underline{83.1} & 32.0 & 37.1 & \underline{32.8} & \underline{39.4} & 18.0 & \underline{76.8} & 8.8 & 48.1 & 20.5 & \underline{60.1} \\
        SegEarth-OV \cite{li2025segearth} & CVPR’25 & \underline{19.3} & 57.9 & ~7.9 & 24.1 & \underline{33.1} & 26.3 & 26.9 & 13.5 & \underline{23.0} & 66.2 & 16.4 & 29.4 & 21.1 & 36.2 \\
        FGAseg \cite{li2025fgaseg} & arXiv’25 & 13.4 & 70.6 & \underline{47.5} & 54.4 & ~7.2 & \underline{37.2} & 23.6 & 38.2 & 13.4 & 71.4 & \underline{18.7} & \underline{55.1} & 20.6 & 54.5 \\
        \textbf{MM-OVSeg} & - & \textbf{19.6} & \textbf{76.7} & \textbf{53.0} & \textbf{86.5} & \textbf{35.2} & \textbf{37.4} & \textbf{40.0} & \textbf{40.4} & \textbf{23.5} & \textbf{77.7} & \textbf{22.7} & \textbf{56.1} & \textbf{32.3} & \textbf{62.5} \\
    \hline
    \end{tabular}}
    \caption{Performance splits for unseen and seen classes. The table reports mIoU scores for each setting and the overall mean.}
\label{tab:performanceSplit}
\end{table*}

\begin{table*}[t] 
    \setlength{\tabcolsep}{10pt}
    \renewcommand{\arraystretch}{1.15}
    \centering
    \begin{tabular}{l|c|c|cccccc|c}
    \toprule
        Method & Backbone & Publication & \ding{172} & \ding{173} & \ding{174} & \ding{175} & \ding{176} & \ding{177} & Mean \\
    \midrule




         CAT-Seg \cite{cho2024catseg} & ViT-L/14 & CVPR'24 & 60.7 & 69.8 & 38.1 & 40.2 & 62.2 & 41.5 & 52.1 \\
         EBSeg \cite{shan2024open} & ViT-L/14 & CVPR'24 & 48.9 & 50.6 & 23.6 & 26.9 & 50.7 & 31.5 & 38.7 \\
        FGAseg \cite{li2025fgaseg} & ViT-L/14 & arXiv'25 & 57.1 & 48.8 & 23.3 & 36.5 & 52.3 & 41.5 & 43.3 \\
        \textbf{MM-OVSeg (ours)} & ViT-L/14 & -- & \textbf{64.5} & \textbf{75.5} & \textbf{39.4} & \textbf{41.2} & \textbf{66.0} & \textbf{43.5} & \textbf{55.0} \\
    \bottomrule
    \end{tabular}
    \caption{Comparison of OVS methods with ViT-L/14 backbone across all evaluation settings  as illustrated in Table \ref{tab:datasets} of the main paper. The table reports mIoU scores for each setting and the overall mean. Settings correspond to: \ding{172}: PIE-cloud→PIE-cloud; \ding{173}: DDHR-SK→DDHR-SK; \ding{174}: OEM-thick→OEM-thick; \ding{175}: OEM-thin→OEM-thin; \ding{176}: PIE-clean→PIE-clean; \ding{177}: DDHR-SK→DDHR-CH. MM-OVSeg achieves the highest accuracy in all settings and obtains the best overall mean score, demonstrating strong robustness under cloudy conditions and superior cross-domain generalization.}
    \label{tab:tablea}
\end{table*}

\begin{figure*}[ht]
\centering
\includegraphics[width=\linewidth]{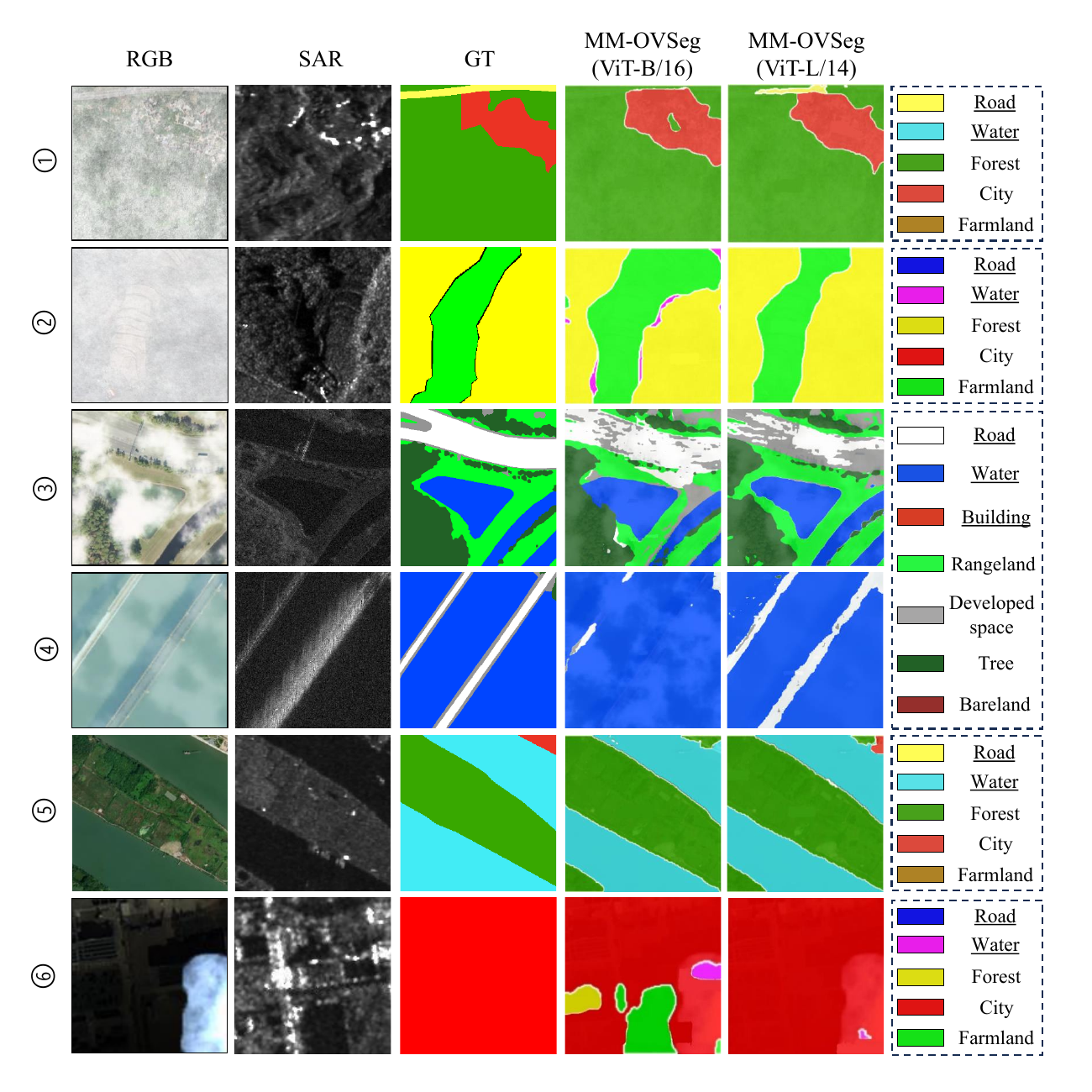}
\caption{Visualization of OVS results. From left to right: input RGB image, input SAR image, ground truth, and segmentation outputs from MM-OVSeg (ViT-B/16) and MM-OVSeg (ViT-L/14). In the legend, underlined categories represent \textit{unseen} classes and the remaining categories are \textit{seen} classes.}
\label{fig:figa3}
\end{figure*}

 \section{Larger Backbones on MM-OVSeg} \label{sec:vitl14}

 We also evaluate how backbone capacity affects the performance of MM-OVSeg.
While the main paper uses ViT-B/16, here we replace both the CLIP and DINO encoders with ViT-L/14, using pretrained weights from CLIP \cite{radford2021learning} and DINO v3 \cite{simeoni2025dinov3}, respectively.
As before, multi-scale features are taken from the 8th, 16th, and 24th transformer blocks.
During full MM-OVSeg training, we train for 120k iterations using AdamW with a batch size of 4 and an initial learning rate of $2.5 \times 10^{-4}$. All other settings follow those used for the ViT-B/16 backbone.

Following the evaluation protocol in Table \ref{tab:table2} of the main paper, Table \ref{tab:tablea} compares MM-OVSeg (ViT-L/14) with recent state-of-the-art OVS methods across all benchmark datasets. Mean IoU is reported for each dataset and the overall average.

Consistent with the trend observed using ViT-B/16, MM-OVSeg (ViT-L/14) achieves the best overall performance, obtaining 55.0\% mIoU across six benchmarks, outperforming the ViT-B/16 version (51.7\%) due to its increased model capacity.
This improvement further validates the strength of our multimodal fusion design for open-vocabulary segmentation in remote sensing. Moreover, MM-OVSeg (ViT-L/14) maintains a substantial lead on setting \ding{177}: DDHR-SK→DDHR-CH, demonstrating strong cross-domain robustness. 

Figure~\ref{fig:figa3} provides visual comparisons between MM-OVSeg (ViT-L/14) and MM-OVSeg (ViT-B/16).
Consistent with the quantitative results in Table~\ref{tab:tablea}, the ViT-L/14 variant produces clearer boundaries, more stable predictions under cloud cover, and more accurate responses on both seen and unseen categories.
This intuitive improvement further demonstrates how increasing model capacity strengthens multimodal fusion, reinforcing the effectiveness of our design for open-vocabulary segmentation in remote sensing.

 
 \section{ CMU for CLIP-SAR Alignment} \label{sec:sarclip}

As discussed in Section \ref{sec:32} of the main paper, we also investigate whether a CLIP-style visual encoder can be trained for SAR using the CMU procedure, analogous to the DINO-SAR setup.
Following the same strategy, we distill multi-scale ViT features from the RGB CLIP encoder into a SAR-specific CLIP encoder using the InfoNCE loss.
Table \ref{tab:distill} reports the performance of four model variants:
\begin{itemize}
    \item Model \#1: baseline without CMU;
    \item Model \#2: CMU applied to CLIP only (CLIP-SAR);
    \item Model \#3: CMU applied to DINO only (DINO-SAR), which corresponds to MM-OVSeg;
    \item Model \#4: CMU applied to both DINO and CLIP for SAR.
\end{itemize}

All CMU variants improve over the baseline, confirming the value of cross-modal alignment.
However, DINO-SAR alone (Model \#3) achieves the best performance, while adding a CLIP-SAR encoder (Model \#4) results in a performance drop. This behavior can be explained as follows: DINO provides dense, locally discriminative features that are crucial for pixel-level segmentation, whereas CLIP encoders produce coarse global embeddings optimized for image-level alignment rather than spatial precision. Training a CLIP-SAR encoder substantially increases the number of global embeddings without providing new local information, which introduces redundancy and complicates the fusion process.

 \begin{table}[t]
    \centering
    \begin{tabular}{c|cc|c}
    \toprule
        Model index & CLIP & DINO & mIoU\\
    \midrule
        \#1 & \xmark & \xmark & 64.1 \\
        \#2 & \cmark & \xmark & 65.4 \\
        \#3 & \xmark & \cmark & \textbf{73.1} \\
        \#4 & \cmark & \cmark & 66.5 \\
    \bottomrule
    \end{tabular} 
    \caption{Ablation on applying CMU to different visual encoders on the \ding{173}: DDHR-SK→DDHR-SK segmentation task.}
    \label{tab:distill}
\end{table}

\section{Additional Visualization Results} \label{sec:visual}

\begin{figure*}[t]
\centering
\includegraphics[width=6in]{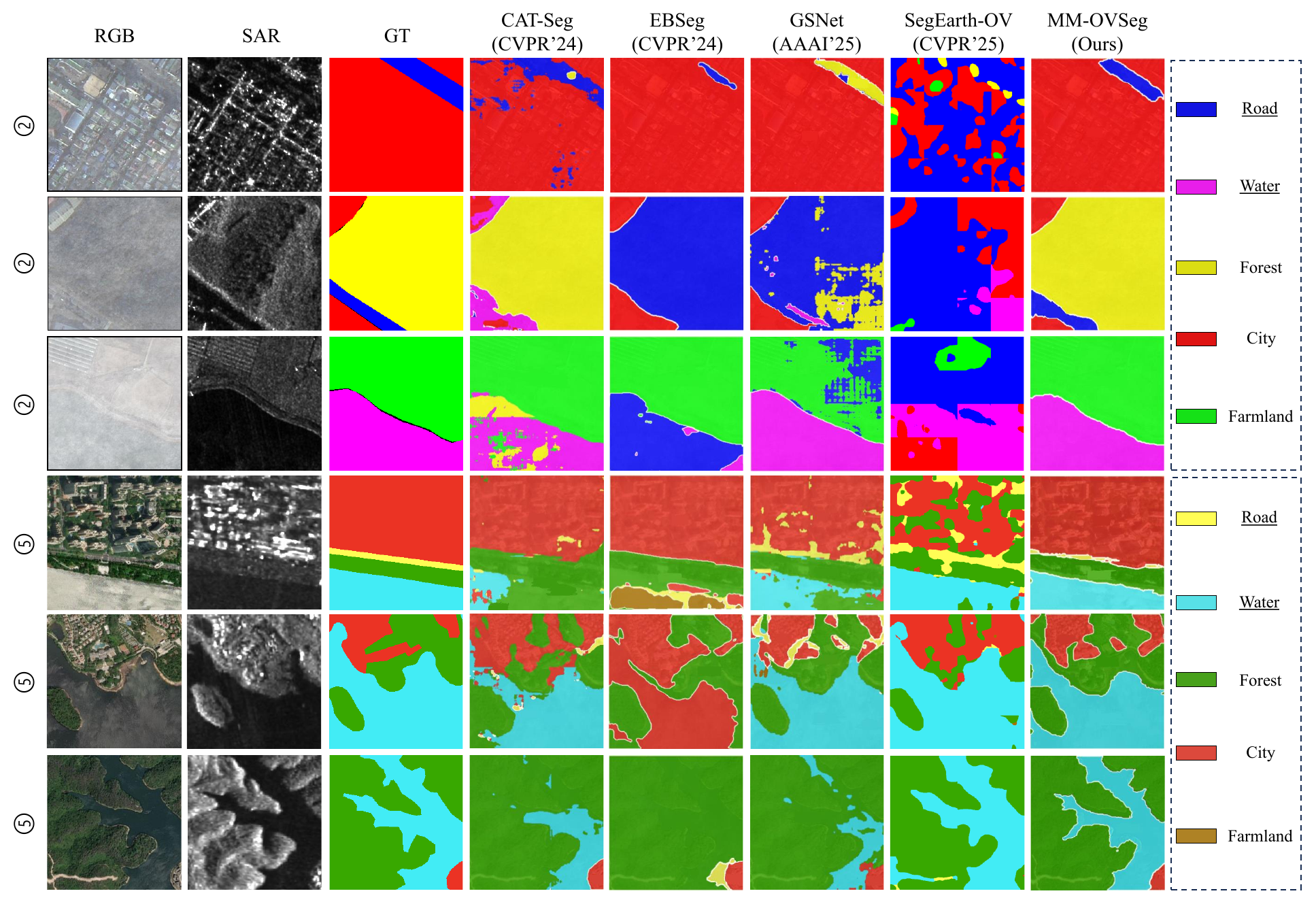} 
\caption{Intra-domain visualization of OVS results, including \ding{173}: DDHR-SK→DDHR-SK and \ding{176}: PIE-clean → PIE-clean. From left to right: input RGB image, input SAR image, ground truth, and segmentation outputs from CAT-Seg, EBSeg, GSNet, SegEarth-OV, and our MM-OVSeg. In the legend, underlined categories represent \textit{unseen} classes and the remaining categories are \textit{seen} classes.}
\label{fig:figa1}
\end{figure*}

\begin{figure*}[t]
\centering
\includegraphics[width=6.2in]{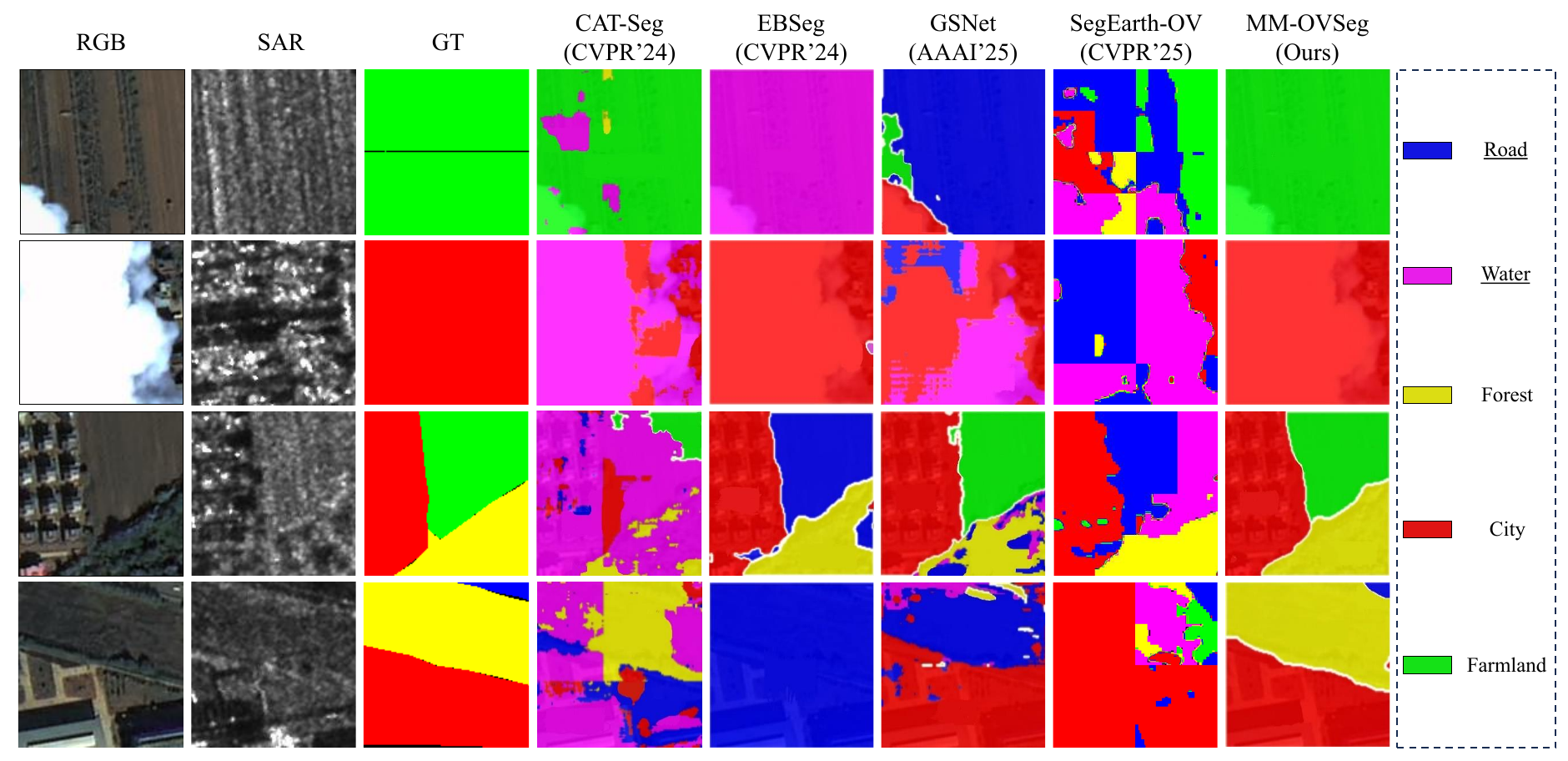}
\caption{Cross-domain visualization of OVS results for \ding{177}: DDHR-SK→DDHR-CH. From left to right: input RGB image, input SAR image, ground truth, and segmentation outputs from CAT-Seg, EBSeg, GSNet, SegEarth-OV, and our MM-OVSeg. In the legend, underlined categories represent \textit{unseen} classes and the remaining categories are \textit{seen} classes.}
\label{fig:figa2}
\end{figure*}

Similar to Figure \ref{fig:fig3} of the main paper, we provide more qualitative comparisons of MM-OVSeg (as in Table \ref{tab:table2} “SOTA performance") for both intra-domain and cross-domain settings. Figure~\ref{fig:figa1} shows additional intra-domain examples, and Figure~\ref{fig:figa2} presents cross-domain results. These results further demonstrate the superiority of MM-OVSeg for multimodal open-vocabulary segmentation across diverse weather conditions.

\end{document}